\documentclass[review]{elsarticle}

\usepackage{hyperref}

\usepackage{natbib}
\usepackage{siunitx}

\journal{Current Opinion in Neurobiology}

\makeatletter
\def\els@aparagraph[#1]#2{\elsparagraph[#1]{#2}}
\def\els@bparagraph#1{\elsparagraph*{#1}}
\makeatother

\begin{document}

\begin{frontmatter}

\title{Convolutional nets for reconstructing neural circuits from brain images acquired by serial section electron microscopy}




\author[mitbcs]{Kisuk Lee\corref{equalcontribution}}
\ead{kisuklee@mit.edu}

\author[princetoncs]{Nicholas Turner\corref{equalcontribution}}
\ead{nturner@cs.princeton.edu}

\author[princetoncs]{Thomas Macrina}
\ead{tmacrina@princeton.edu}

\author[princetonpni]{Jingpeng Wu}
\ead{jingpeng@princeton.edu}

\author[princetonpni]{Ran Lu}
\ead{ranl@princeton.edu}

\author[princetoncs,princetonpni]{H. Sebastian Seung\corref{correspondingauthor}}
\ead{sseung@princeton.edu}

\cortext[equalcontribution]{Equal contribution}
\cortext[correspondingauthor]{Corresponding author}
\address[mitbcs]{Department of Brain and Cognitive Sciences, MIT, Cambridge, MA 02139, USA}
\address[princetoncs]{Department of Computer Science, Princeton University, Princeton, NJ 08544, USA}
\address[princetonpni]{Neuroscience Institute, Princeton University, Princeton, NJ 08544, USA}

\begin{abstract}
Neural circuits can be reconstructed from brain images acquired by serial section electron microscopy. Image analysis has been performed by manual labor for half a century, and efforts at automation date back almost as far. Convolutional nets were first applied to neuronal boundary detection a dozen years ago, and have now achieved impressive accuracy on clean images. Robust handling of image defects is a major outstanding challenge. Convolutional nets are also being employed for other tasks in neural circuit reconstruction: finding synapses and identifying synaptic partners, extending or pruning neuronal reconstructions, and aligning serial section images to create a 3D image stack. Computational systems are being engineered to handle petavoxel images of cubic millimeter brain volumes.
\end{abstract}


\end{frontmatter}


\section*{Introduction}
The reconstruction of the \emph{C. elegans} nervous system by serial section electron microscopy (ssEM) required years of laborious manual image analysis \citep{white1986structure}. Even recent ssEM reconstructions of neural circuits have required tens of thousands of hours of manual labor \citep{lee2016anatomy}. 
The dream of automating EM image analysis dates back to the dawn of computer vision in the 1960s and 70s (Marvin Minsky, personal communication) \citep{sobel1980special}. In the 2000s, connectomics was one of the first applications of convolutional nets to dense image prediction \citep{jain2007supervised}. More recently, convolutional nets were finally accepted by mainstream computer vision, and enhanced by huge investments in hardware and software for deep learning. It now seems likely that the dream of connectomes with minimal human labor will eventually be realized with the aid of artificial intelligence (AI). 

More specific impacts of AI on the technologies of connectomics are harder to predict. One example is the ongoing duel between 3D EM imaging approaches, serial section and block face \citep{kornfeld2018progress}. Images acquired by ssEM may contain many defects, such as damaged sections and misalignments, and axial resolution is poor. Block face EM (bfEM) was originally introduced to deal with these problems \citep{denk2004serial}. For its fly connectome project, Janelia Research Campus has invested heavily in FIB-SEM \citep{xu2017enhanced}, a bfEM technique that delivers images with isotropic 8~nm voxels and few defects \citep{knott2008serial}. FIB-SEM quality is expected to boost the accuracy of image analysis by AI, thereby reducing costly manual image analysis by humans. Janelia has decided that this is overall the cheapest route to the fly connectome, even if FIB-SEM imaging is expensive (Gerry Rubin, personal communication).

It is possible that the entire field of connectomics will eventually switch to bfEM, following the lead of Janelia. But it is also possible that powerful AI plus lower quality ssEM images might be sufficient for delivering the accuracy that most neuroscientists need. The future of ssEM depends on this possibility. 

The question of whether to invest in better data or better algorithms often arises in AI research and development. For example, most self-driving cars currently employ an expensive LIDAR sensor with primitive navigation algorithms, but cheap video cameras with more advanced AI may turn out to be sufficient in the long run \citep{bojarski2016end}. 

AI is now highly accurate at analyzing ssEM images under ``good conditions,'' and continues to improve.  But AI can fail catastrophically at image defects. This is currently its major shortcoming relative to human intelligence, and the toughest barrier to creating practical AI systems for ssEM images. The challenge of robustly handling edge cases is universal to building real-world AI systems, and makes the difference between a self-driving car for research and one that will be commercially successful.

\begin{figure}[t!]
\begin{center}
\includegraphics[width=1.0\textwidth]{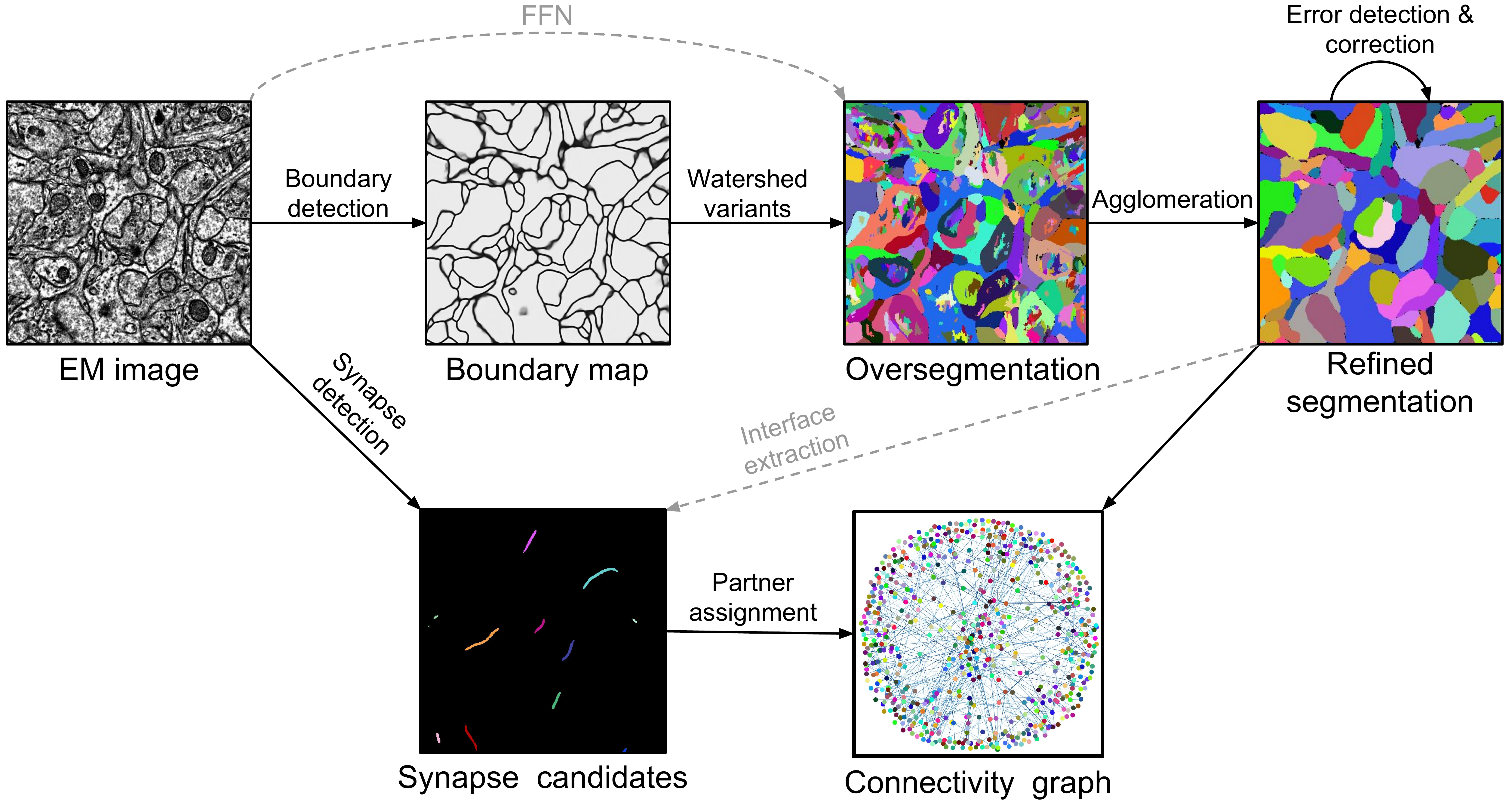}
\end{center}
\caption{Example of connectomic image processing system. Neuronal boundaries and synaptic clefts are detected by convolutional nets. The boundary map is oversegmented into supervoxels by a watershed-type algorithm. Supervoxels are agglomerated to produce segments using hand-designed heuristics or machine learning algorithms. Synaptic partners in the segmentation are assigned to synapses using convolutional nets or other machine learning algorithms. Error detection and correction is performed by convolutional nets or human proofreaders. Gray dashed lines indicate alternative paths taken by some methods. 
\label{fig:pipeline}}
\end{figure}

In this journal, our lab previously reviewed the application of convolutional nets to connectomics \citep{jain2010machines}. The ideas were first put into practice in a system that applied a single convolutional net to convert manually traced neuronal skeletons into 3D reconstructions \citep{helmstaedter2013inner}, and a semiautomated system (Eyewire) in which humans reconstructed neurons by interacting with a similar convolutional net \citep{kim2014space-time}.\footnote{Other early semiautomated reconstruction systems did not use convolutional nets \citep{mishchenko2010ultrastructural, takemura2013visual}.}  A modern connectomics platform (Fig.~\ref{fig:pipeline}) now contains multiple convolutional nets performing a variety of image analysis tasks. 

This review will focus on ssEM image analysis. At present ssEM remains the most widely used technique for neural circuit reconstruction, judging from number of publications \citep{kornfeld2018progress}. The question for the future is whether AI will be able to robustly handle the deficiencies of ssEM images. This will likely determine whether ssEM will remain popular, or be eclipsed by bfEM. 

Serial section EM was originally done using transmission electron microscopy (TEM). More recently, serial sections have also been imaged by scanning electron microscopy (SEM) \citep{kasthuri2015saturated}. Serial section TEM and SEM are more similar to each other, and more different from bfEM techniques. Both ssEM methods produce images that can be reconstructed by very similar algorithms, and we will typically not distinguish between them in the following. 


\section*{Alignment}


The connectomic image processing system of Fig.~\ref{fig:pipeline} accepts as input a 3D image stack. This must be assembled from many raw ssEM image tiles. It is relatively easy to stitch or montage multiple image tiles to create a larger 2D image of a single section. Adjacent tiles contain the same content in the overlap region, and mismatch due to lens distortion can be corrected fairly easily \citep{kaynig2015large}.  More challenging is the alignment of 2D images from a series of sections to create the 3D image stack. Images from serial sections actually contain different content, and physical distortions of sections can cause severe mismatch.

The classic literature on image alignment has included two major approaches. One is to find corresponding points between image pairs, and then compute transformations of the images that bring corresponding points close together. In the iterative rerendering approach, one defines an objective function such as mean squared error or mutual information after alignment, and then iteratively searches for the parametrized image transformations that optimize the objective function. The latter approach is popular in human brain imaging \citep{avants2011reproducible}. It has not been popular for ssEM images, possibly because iterative rerendering is computationally costly.

The corresponding points approach has been taken by a number of ssEM software packages, including TrakEM2 \citep{saalfeld2012elastic}, AlignTK \citep{wetzel2016registering}, NCR Tools \citep{anderson2011exploring}, FijiBento \citep{joesch2016reconstruction}, and EM\_aligner \citep{khairy2018joint}. As an example of the state-of-the-art, it is helpful to examine a recent whole fly brain dataset \citep{zheng2018complete}, which is publicly available. The alignment appears outstanding at most locations. Large misalignments do occur rarely, however, and small misalignments can also be seen. This level of quality is sufficient for manual reconstruction, as humans are smart enough to compensate for misalignments. However, it poses challenges for automated reconstruction. 

To improve alignment quality, several directions are being explored. One can add extensive human-in-the-loop capabilities to traditional computer vision algorithms, enabling a skilled operator to adjust parameters on a section-by-section basis, as well as detect and remove false correspondences \citep{macrina2018alembic} . Another approach is to reduce false correspondences from normalized cross correlation using weakly supervised learning of image encodings \citep{buniatyan2017deep}. 


The iterative rerendering approach has been extended by \citet{yoo2017deep}, who define an objective function based on the difference between image encodings rather than images. The encodings are generated by unsupervised learning of a convolutional autoencoder.

Another idea is to train a convolutional net to solve the alignment problem, with no optimization at all required at run-time.  For example, one can train a convolutional net to take an image stack as input, and return an aligned image stack as output \citep{jain2017adversarial}. Alternatively, one can train a convolutional net to take an image stack as input, and return deformations that align it \citep{mitchell2019siamese}. Similar approaches are already standard in optical flow \citep{dosovitskiy2015flownet,ranjan2017spynet}.

\begin{figure}[t!]
\includegraphics[width=\textwidth]{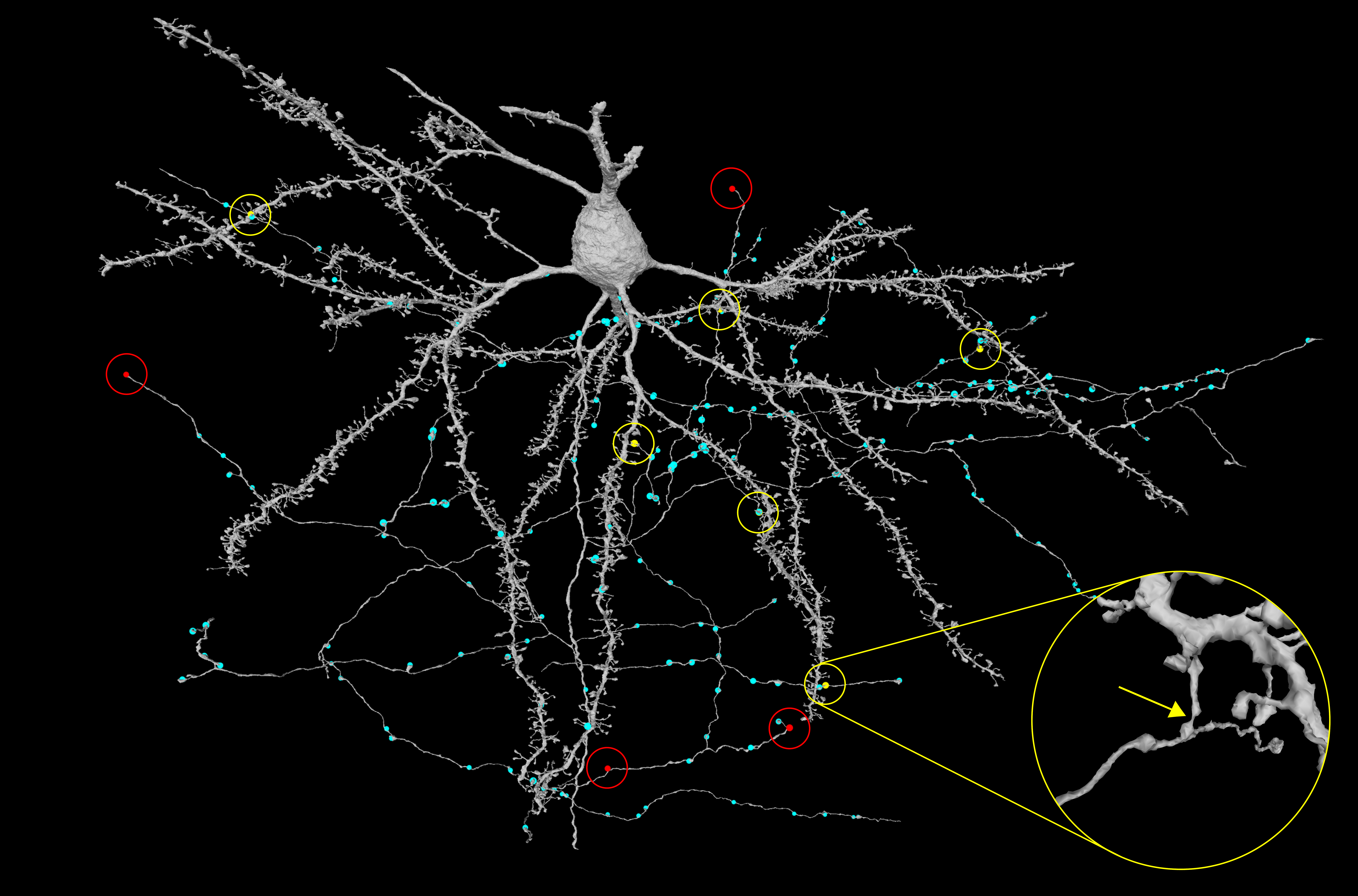}
\begin{center}
\end{center}\caption{Automated reconstruction of pyramidal neuron in mouse visual cortex by a system like that of Fig.~\ref{fig:pipeline}. The reconstruction is largely correct, though one can readily find split errors (red, 4) and merge errors (yellow, 6) in proofreading. 6 of these errors are related to image artifacts, and merge errors often join an axon to a dendritic spine. Cyan dots represent automated predictions of presynaptic sites. Predicted postsynaptic terminals were omitted for clarity.  The $200\times 100\times 100$ $\mu$m$^3$ ssEM dataset was acquired by the Allen Institute for Brain Science (A. Bodor, A. Bleckert, D. Bumbarger, N. M. da Costa, C. Reid) after calcium imaging \emph{in vivo} by the Baylor College of Medicine (E. Froudarakis, J. Reimer,  Andreas S. Tolias). Neurons were reconstructed by Princeton University (D. Ih, C. Jordan, N. Kemnitz, K. Lee, R. Lu, T. Macrina, S. Popovych, W. Silversmith, I. Tartavull, N. Turner, W. Wong, J. Wu, J. Zung,  and H. S. Seung). 
}\label{fig:ExampleReconstruction}
\end{figure}

\section*{Boundary detection}

Once images are aligned into a 3D image stack, the next step is \textit{neuronal boundary detection} by a convolutional net (Fig. \ref{fig:pipeline}). In principle, if the resulting \textit{boundary map} were perfectly accurate, then it would be trivial to obtain a segmentation of the image into neurons ~\citep{jain2007supervised,jain2010machines}.


Detecting neuronal boundaries is challenging for several reasons. First, many other boundaries are visible inside neurons, due to intracellular organelles such as mitochondria and endoplasmic reticulum. Second, neuronal boundaries may fade out in locations, due to imperfect staining. Third, neuronal morphologies are highly complex. The densely packed, intertwined branches of neurons make for one of the most challenging biomedical image segmentation problems.

A dozen years ago, convolutional nets were already shown to outperform traditional image segmentation algorithms at neuronal boundary detection~\citep{jain2007supervised}. Since then advances in deep learning have dramatically improved boundary detection accuracy \citep{zeng2017deepem3d,beier2017multicut,lee2017superhuman,funke2018structured}, as evidenced by two publicly available challenges on ssEM images, SNEMI3D\footnote{SNEMI3D challenge; URL: \url{http://brainiac2.mit.edu/SNEMI3D/}}~ and CREMI.\footnote{CREMI challenge; URL: \url{https://cremi.org/}}

How do state-of-the-art boundary detectors differ from a dozen years ago?   \citet{jain2007supervised} used a net with six convolutional layers, eight feature maps per hidden layer, and 34,041 trainable parameters. \citet{funke2018structured} use a convolutional net containing pathways as long as 18 layers, as many as 1,500 feature maps per layer, and 83,998,872 trainable parameters.

State-of-the-art boundary detectors use the U-Net architecture \citep{ronneberger2015unet,cicek2016unet} or variants \citep{zeng2017deepem3d, lee2017superhuman, funke2018structured, quan2016fusionnet, fakhry2017resdeconv}. The multiscale architecture of the U-Net is well-suited for handling both small and large neuronal objects, i.e., detecting boundaries of thin axons and spine necks, as well as thick dendrites. An example of automated reconstruction with a U-Net style architecture is shown in Fig. \ref{fig:ExampleReconstruction}.

State-of-the-art boundary detectors make use of 3D convolutions, either exclusively or in combination with 2D convolutions. Already a dozen years ago, the first convolutional net applied to neuronal boundary detection used exclusively 3D convolutions~\citep{jain2007supervised}. However, this net was applied to bfEM images, which have roughly isotropic voxels and are usually well-aligned. It was popular to think that 3D convolutional nets were not appropriate for ssEM images, which had much poorer axial resolution and suffered from frequent misalignments. Furthermore, many deep learning frameworks did not support 3D convolutions very well or at all. Accordingly, much work on ssEM images has used 2D convolutional nets \citep{knowles-barley2016rhoananet, meirovitch2016multipath}, relying on agglomeration techniques to link up superpixels in the axial direction. Today it has become commonly accepted that 3D convolutions are also useful for processing ssEM images.



The current SNEMI3D \citep{lee2017superhuman} and CREMI leaders \citep{beier2017multicut,funke2018structured} both generate nearest neighbor \textit{affinity graphs} as representations of neuronal boundaries. The affinity representation was introduced by \citet{turaga2010affinity} as an alternative to classification of voxels as boundary or non-boundary. It is especially helpful for representing boundaries in spite of the poor axial resolution of ssEM images~\citep{jain2010machines}. \citet{parag2017aniso} have published empirical evidence that the affinity representation is helpful.

\citet{turaga2009malis} pointed out that using the Rand error as a loss function for boundary detection would properly penalize topological errors, and proposed the MALIS method for doing this. \citet{funke2018structured} have improved upon the original method through constrained MALIS training. \citet{lee2017superhuman} used long-range affinity prediction as an auxiliary objective during training, where prediction error for the affinities of a subset of long-range edges can be viewed as a crude approximation to the Rand error.

\section*{Handling of image defects}
The top four entries on the SNEMI3D leaderboard have surpassed human accuracy as previously estimated by disagreement between two humans. But the claim of ``superhuman'' performance  comes with many caveats \citep{lee2017superhuman}. The chief one is that the SNEMI3D dataset is relatively free of image defects. Robust handling of image defects is crucial for real-world accuracy, and is where human experts outshine AI. 


One can imagine several ways of handling image defects: (1) Heal the defect by some image restoration computation. (2) Make the boundary detector more robust to the defect. (3) Correct the output of the boundary detector by subsequent processing steps. Below, we detail different types of image defects, their effects on reconstruction, and efforts to account for them.

\paragraph*{Missing sections} are not infrequent in ssEM images. Entire sections can be lost during cutting, collection, or imaging. The rate of loss varies across datasets, e.g., 0.17\% in~\citet{zheng2018complete} and 7.56\% in~\citet{tobin2017variations}. It is also common to lose part of a section. For example, one might exclude from imaging the regions of sections that are damaged or contaminated. One might also throw out image tiles after imaging, if they are inferior in quality. Or the imaging system might accidentally fail on certain tiles.

In an ssEM image stack, a partially missing section typically appears as an image with a black region where data is missing. An entirely missing section might be represented by an image that is all black, or might be omitted from the stack.

Traceability of neurites by human or machine is typically high if only a single section is missing. Traceability drops precipitously with long stretches of consecutive missing sections.\footnote{The longest stretch of missing sections was six in~\citet{tobin2017variations} and eight in~\citet{lee2016anatomy}, which amounts to the loss of about \SI{300}{\nano\metre}-thick tissue.} 

Partially and entirely missing sections are easy to simulate during training; one simply blacks out part or all of an image. \citet{funke2018structured} simulated entirely missing sections during training at a 5\% rate. \citet{lee2017superhuman} simulated both partially and entirely missing sections at a higher rate, and found that convolutional nets can learn to ``imagine'' an accurate boundary map even with several consecutive missing sections.


\paragraph*{Misalignments} are not image defects, strictly speaking. They arise during image processing, after image acquisition. From the perspective of the boundary detector, however, misalignments appear as input defects.
Progress in alignment software and algorithms is rapidly reducing the rate of misalignment errors. Nevertheless, misalignments can be the dominant cause of tracing errors, because boundary detection in the absence of image defects has become so accurate. 

\citet{lee2017superhuman} introduced a novel type of data augmentation for simulating misalignments during training. Injecting simulated misalignments at a rate much higher than the real rate was shown to improve the robustness of boundary detection to misalignment errors. 

\citet{januszewski2018high-precision} locally realigned image subvolumes prior to agglomeration, in an attempt to remove the image defect before further image processing.

\paragraph*{Cracks and folds} are common in ssEM sections. They may involve true loss of information, or cause misalignments in neighboring areas. We expect that improvements in software will be able to correct misalignments neighboring cracks or folds. 

\paragraph*{Knife marks} are linear defects in ssEM images caused by imperfect cutting at one location on the knife blade. They can be seen in publicly available ssEM datasets \citep{kasthuri2015saturated}.\footnote{\url{https://neurodata.io/data/kasthuri15/}} They are particularly harmful because they occur repeatedly at the same location in consecutive serial sections, and are difficult to simulate. Even human experts have difficulty tracing through knife marks. 




\section*{Agglomeration}

There are various approaches to generate a segmentation from the output of the boundary detector. The na\"ive approach of thresholding the boundary map and computing connected components can lead to many merge errors caused by ``noisy'' prediction of boundaries. Instead, it is common to first generate an oversegmentation into many small supervoxels. Watershed-type algorithms can be used for this step \citep{zlateski2015watershed}. The number of watershed domains can be reduced by size-dependent clustering \citep{zlateski2015watershed}, seeded watershed combined with distance transformation~\citep{beier2017multicut,funke2018structured}, or machine learning~\citep{wolf2017learned}.

Supervoxels are then agglomerated by various approaches. A classic computer vision approach is to use statistical properties of the boundary map~\citep{arbelaez2011contour}, such as mean affinity \citep{lee2017superhuman} or percentiles of binned affinity \citep{funke2018structured}. A score function can be defined for every pair of contacting segments. At every step, the pair with the highest score is merged. This simple procedure can yield surprisingly accurate segmentations when starting from high quality boundary maps, and can be made computationally efficient for large scale segmentation.

Agglomeration can also utilize other information not contained in the boundary map, such as features extracted from the input images or the segments~\citep{jain2011aggl,bogovic2013aggl,nunez-iglesias2013gala,nunez-iglesias2014gala}. Machine learning methods can also be used directly without defining underlying features to serve as the scoring function to be used in the agglomeration iterations~\citep{maitin-shepard2016celis}. 

Supervoxels can also be grouped into segments by optimization of global objective functions \citep{beier2017multicut}. Success of this approach depends on designing a good objective function and algorithms for approximate solution of the typically NP-hard optimization problem.

\section*{Error detection and correction}
Convolutional nets are also being used to automatically detect errors in neuronal segmentations \citep{haehn2018guided,rolnick2017morphological,dmitriev2018efficient,zung2017error}. 
\citet{dmitriev2018efficient} leverage skeletonization of candidate segments, applying convolutional nets selectively to skeleton joints and endpoints to detect merge and split errors, respectively. \citet{rolnick2017morphological} train a convolutional net to detect artificially induced merge errors. \citet{zung2017error} demonstrate detection and localization of both split and merge errors with supervised learning of multiscale convolutional nets.

\begin{figure}[t!]
\begin{center}
\includegraphics[width=1.0\textwidth]{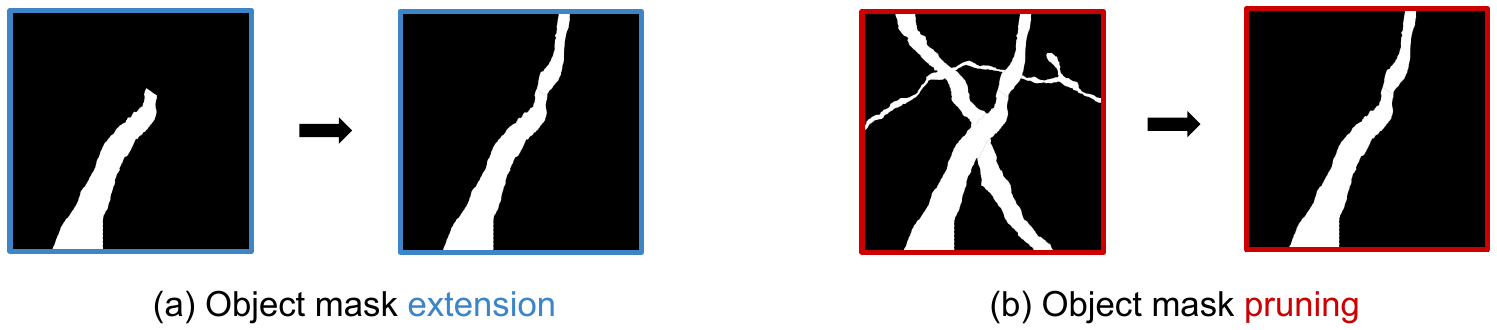}
\end{center}
\caption{Illustration of (a) object mask extension~\citep{meirovitch2016multipath,januszewski2018high-precision} and (b) object mask pruning~\citep{zung2017error}. Both employ attentional mechanisms, focusing on one object at a time. Object mask extension takes as input a subset of the true object and adds missing parts, whereas object mask pruning takes as input a superset of the true object and subtracts excessive parts. Both tasks typically use the raw image (or some alternative representation) as an extra input to figure out the right answer, though object mask pruning may be less dependent on the raw image~\citep{zung2017error}.
\label{fig:error}}
\end{figure}

Convolutional nets have also been used to correct morphological reconstruction errors \citep{meirovitch2016multipath,haehn2018guided,dmitriev2018efficient,zung2017error}. \citet{zung2017error} propose an error-correcting module which prunes an ``advice'' object mask constructed by aggregating erroneous objects found by an error-detecting module. The ``object mask pruning'' task (Fig.~\ref{fig:error}b) is an interesting counterpoint to the ``object mask extension'' task implemented by other methods (Fig.~\ref{fig:error}a) \cite{januszewski2018high-precision}.

\section*{Synaptic relationships}

To map a connectome, one must not only reconstruct neurons, but also determine the synaptic relationships between them. The annotation of synapses has traditionally been done manually, yet this is infeasible for larger volumes \cite{dorkenwald2017}. 
Most research on automation has focused on chemical synapses. This is because large volumes are typically imaged with a lateral resolution of 4~nm or worse, which is insufficient for visualizing electrical synapses.  Higher resolution would increase both imaging time and dataset size.

A central operation in many approaches is the classification of each voxel as ``synaptic'' or ``non-synaptic'' \citep{kreshuk2011,becker2013,kreshuk2014,jagadeesh2014,marquezneila2016}. 
Increasingly, convolutional nets are being used for this voxel classification task \citep{roncal2014,huang2014,huang2016,dorkenwald2017,santurkar2017,heinrich2018cleft,buhmann2018partner}. Synaptic clefts are then predicted by some hand-designed grouping of synaptic voxels.


A number of approaches are used to find the partner neurons at a synaptic cleft \citep{roncal2014,huang2016,dorkenwald2017,santurkar2017,parag2018detecting}. For example, \citet{dorkenwald2017} extract features relating the predicted cleft segments to candidate partners by overlap with these partners, as well as their contact site, and feed these features to a random forest classifier for a final ``synaptic'' or ``non-synaptic'' classification for inferring connectivity. \citet{parag2018detecting} pass a candidate cleft, local image context, and a candidate pair of partner segments to a convolutional net to make a similar ``synaptic'' or ``non-synaptic'' judgment. \citet{turner2019synaptic} use a similar model, yet they instead use the cleft as an attentional input to predict the voxels of relevant presynaptic and postsynaptic partners.

Another group of recent approaches do not explicitly represent synaptic clefts ~\citep{buhmann2018partner,staffler2017}. \citet{buhmann2018partner} detect sites where presynaptic and postsynaptic terminals are separated by specific spatial intervals, using a single convolutional net to predict both location and directionality of synapses. \citet{staffler2017} extract contact sites for all adjacent pairs of segments within the region of interest, and use a decision tree trained on hand-designed features of the ``subsegments'' close to the contact site in order to infer synaptic contact sites and their directionality of connection. 


\section*{Distributed chunk-wise processing of large images}
In connectomics, it is often necessary to transform an input image into an output image (Fig. \ref{fig:pipeline}). The large input image is divided into overlapping chunks, each of which is small enough to fit into RAM of a single worker (CPU node or GPU). Each chunk is read from persistent storage by a worker, which processes the chunk and writes the result back to persistent storage. Additional operations may be required to insure consistency in the regions of overlap between the chunks. The processing of the chunks is distributed over many simultaneous workers for speedup.

To implement the above scheme, we need a ``chunk workflow engine'' for distributing chunk processing tasks over workers, and handling dependencies between tasks. We also need a ``cutout service'' that enables workers to ``cut'' chunks out of the large image. This includes both read and write operations, which may be requested by many workers in parallel. The engine and cutout service may function with a local cluster and network-attached storage, or with cloud computing and storage.


\begin{figure}[t!]
\begin{center}
\includegraphics[width=0.8\textwidth]{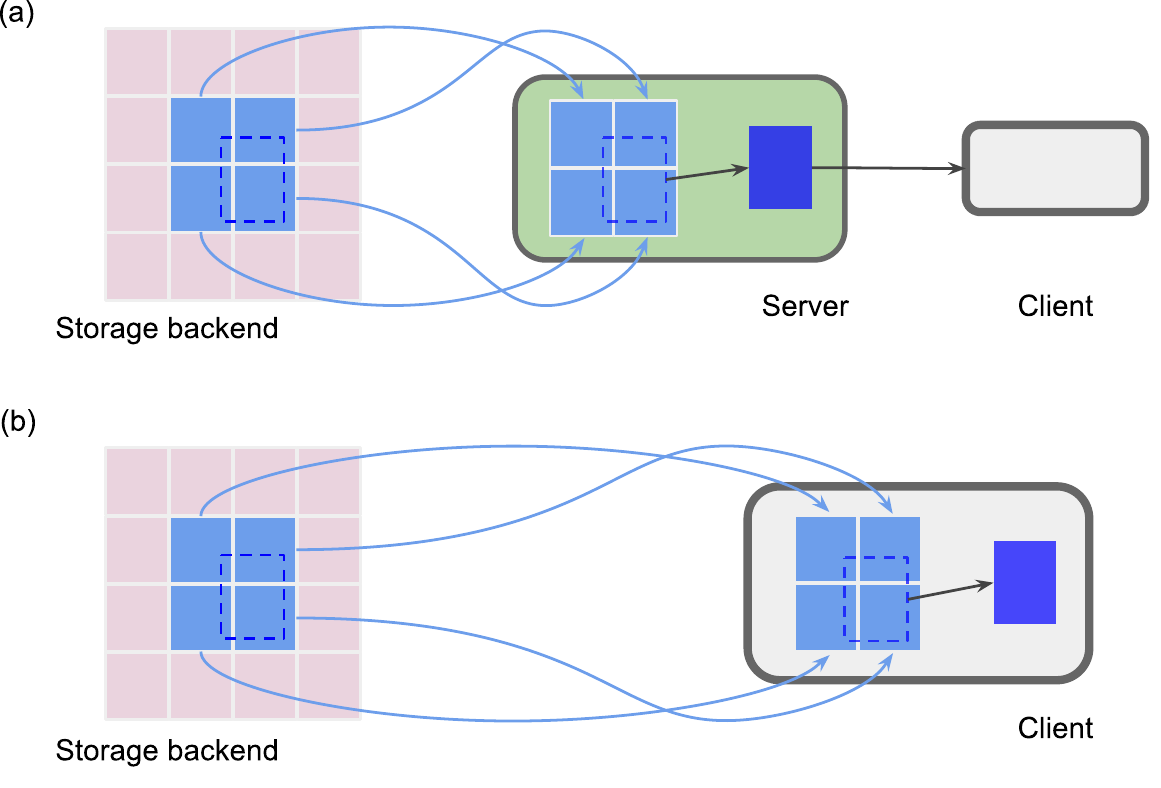}
\end{center}\caption{Two approaches to chunk cutout services. (a) Server-side cutout. A server responds to a cutout request by loading the required image blocks (light blue), performing the cutout (blue), and sending the result back to a client. (b) Client-side cutout. A client requests the chunk blocks from cloud storage, and performs the cutout. A read operation is shown here; write operations are also supported by cutout services.}
\label{fig:cutout}
\end{figure}



DVID is a cutout service used with both network-attached and Google Cloud Storage~\citep{katz2019dvid}. Requests are handled by a server running on a compute node in the local cluster or cloud. For ease of use, DICED is a Python wrapper that makes a DVID store behave as if it were a NumPy array~\citep{plaza2018diced}. bossDB uses microservices (AWS Lambda) to fulfill cutout requests from an image in Amazon S3 cloud storage~\citep{kleissas2017boss}. ndstore shares common roots with bossDB~\citep{burns2013open}, but uses Amazon EC2 servers rather than microservices. It is employed by neurodata.io, formerly known as the Open Connectome Project.

For cloud storage, the cutout service can also be implemented on the client side (Fig.~\ref{fig:cutout}). CloudVolume~\citep{silversmith2018cloudvolume} and BigArrays.jl \citep{wu2018bigarrays} convert cutout requests into native cloud storage requests. Both packages provide convenient slicing interfaces for Python NumPy arrays~\citep{silversmith2018cloudvolume} and Julia arrays~\citep{wu2018bigarrays} respectively.

The chunk workflow engine handles task scheduling, the assigning of tasks to workers. In the simplest case, processing the large image can be decomposed into chunk processing tasks that do not depend on each other at all. Then scheduling can be handled by a task queue. Tasks are added to the queue in arbitrary order. When a worker is idle, it takes the first available task from the queue. The workers continue until the queue is empty. The preceding can be implemented in the cloud using AWS Simple Queue Service \citep{wu2019chunkflow, silversmith2018igneous}. Task scheduling is more complex when there are dependencies between tasks. If workflow dependencies are represented by a directed acyclic graph (DAG), the workflow can be executed by, for example,
Apache Airflow~\citep{wong2018seuron}. 

In large scale distributed processing, errors are almost inevitable. The most common error is that a worker will fail to complete a task. Without proper handling of this failure mode, chunks will end up missing from the output image. A workflow engine typically judges a task to have failed if it is not completed within some fixed waiting time. The task is then reassigned to another worker. This ``retry after timeout'' strategy can be implemented in a variety of ways. In the case of a task queue, when one worker starts a task, the task is made invisible to other workers for a fixed waiting time. If the worker completes the task in time, it deletes the task from the queue. If the worker fails to complete the task in time, the task becomes visible in the queue again, and another worker can take it.

One can reduce cloud computing costs by using unstable instances (called ``preemptible'' on Google and ``spot'' on Amazon). These are much cheaper than on-demand instances, but can be killed at any time in favor of customers using on-demand instances or bidding with a higher price. When an instance is killed, its tasks will fail to complete. Therefore, the error handling described above is not only important for insuring correct results, but also for lowering cost.

\section*{Beyond boundary detection?}
\label{section:beyond}
The idea of segmenting a nearest neighbor affinity graph generated by a convolutional net that detects neuronal boundaries is now standard (Fig. \ref{fig:pipeline}). 
One intriguingly different approach is the flood-filling network (FFN) of \citet{januszewski2018high-precision}. The convolutional net receives the EM image as one input, and a binary mask representing part of one object as an ``attentional'' input. The net is trained to extend the binary mask to cover the entire object within some spatial window (Fig. \ref{fig:error}a). Iteration of the ``flood filling'' operation can be used to trace an object that is much larger than the spatial window. ``Flood filling,'' also called ``object mask extension'' in the computer vision literature, directly reconstructs an object. This is a beautiful simplification, because it eliminates postprocessing by a segmentation algorithm (e.g. connected components or watershed) as required by boundary detection.

That being said, whether FFNs deliver accuracy superior to state-of-the-art boundary detectors is unclear. On the FIB-25 dataset~\citep{takemura2015synaptic}, FFNs \citep{januszewski2018high-precision} outperformed the U-Net boundary detector of \citet{funke2018structured}, but this U-Net may no longer be state-of-the-art according to the CREMI leaderboard. In the SNEMI3D challenge, the accuracies of the FFN and boundary detector approaches are merely comparable \citep{januszewski2018high-precision}.

Currently FFNs have high computational cost relative to boundary detectors, because every location in the image must be covered so many times by the FFN~\citep{januszewski2018high-precision}. Nevertheless, one can imagine that FFNs could become much less costly in the future, riding the tailwinds of huge industrial investment in efficient convolutional net inference. Boundary detectors will at the same time become less costly, but watershed-type postprocessing of boundary maps may not enjoy similar cost reduction.

Another intriguing direction is the application of deep metric learning to neuron reconstruction. As noted above, \citet{lee2017superhuman} have shown that prediction of nearest neighbor affinities can be made more accurate by also training the network to predict a small subset of long-range affinities. For predicting many or all long-range affinities, it is more efficient for a convolutional net to assign a feature vector to every voxel, and compute the affinity between two voxels as a function of their feature vectors. Voxels within the same object receive similar feature vectors and voxels from different objects receive dissimilar feature vectors. This idea was first explored in connectomics by \citet{zung2017error} in an error detection and correction system.

Given the representation from deep metric learning, it is tempting to reconstruct an object by finding all feature vectors that are close in embedding space to a seed vector. This approach has been followed by the computer vision literature \citep{fathi2017metric,brabandere2017discriminative}, effectively portraying deep metric learning as a ``kill watershed'' approach like FFNs. However, seed-based segmentation turns out to be outperformed by segmenting a nearest-neighbor affinity graph generated from the feature vectors \citep{luther2019learning}. Seed-based segmentation often fails on large objects, because the convolutional net cannot learn to generate sufficiently uniform feature vectors inside them.

Whether novel ``object-centered'' approaches like FFNs and deep metric learning can render boundary detectors obsolete remains to be seen. The showdown ensures suspense in the future of AI for connectomics.


\section*{Conclusion}
The automation of ssEM image analysis is important because ssEM is still the dominant approach to reconstructing neural circuits \citep{kornfeld2018progress}. It is also highly challenging for AI because ssEM image quality is poor relative to FIB-SEM. The field has made dramatic progress in increasing accuracy, but human effort is still required to correct the remaining errors of AI. Due to lack of space, we are unable to describe the considerable engineering effort currently going into semi-automated platforms that enable human proofreading of automated reconstructions. These platforms must adapt to leverage the increasing power of AI. As the error rate of automated reconstruction decreases, it may become more time-consuming for the human expert to both detect and correct errors. To make efficient use of human effort, the user interface must be designed and implemented with care.  Automated error detection may be useful for focusing human effort on particular locations~\citep{haehn2018guided}, and automated error correction may be useful for suggesting interventions.

In spite of the gains from AI, the overall amount of human effort employed by connectomics may increase because the volumes being reconstructed are also increasing. Even a cubic millimeter has never been fully reconstructed yet, and a human brain is larger by six orders of magnitude. The demand for neural circuit reconstructions should also grow dramatically as they become less costly. Therefore AI may actually increase the total amount of human effort devoted to connectomics, at least in the short term, bucking the conventional wisdom that AI is a jobs destroyer.

Automation of image analysis is spurring interest in increasing the throughput of image acquisition.  When image analysis was the bottleneck, increasing image quality was perhaps more important. Now that the bottleneck is being relieved, there will be increased demand for high throughput SEM and TEM image acquisition systems \citep{zheng2018complete,eberle2015high}. With continued advances in both image acquisition and analysis, it seems likely that ssEM reconstruction of cubic millimeter volumes will eventually become routine. It is unclear, however, whether ssEM image acquisition can scale up to cubic centimeter volumes (whole mouse brain) or larger. New bfEM approaches are being proposed for acquiring such exascale datasets \citep{hayworth2019gcib}. The battle between ssEM and bfEM continues.


\bibliography{mybibfile}

\end{document}